\documentclass[11pt]{article}
\usepackage{acl2016}

\aclfinalcopy 
\def\aclpaperid{265} 

\usepackage[leqno, fleqn]{amsmath}
\usepackage[breaklinks, colorlinks, linkcolor=black, urlcolor=black, citecolor=black, draft]{hyperref}
\usepackage{natbib}
\usepackage{times}
\usepackage{txfonts}
\usepackage{latexsym}

\usepackage{url}
\usepackage{qtree}
\usepackage{booktabs}
\usepackage{caption}
\usepackage{subcaption}
\usepackage{color}
\usepackage{tikz}
\usepackage{tikz-qtree}
\usepackage{algorithm}
\usepackage[noend]{algpseudocode}
\usepackage{stmaryrd}
\usepackage{gb4e}

\newcount\colveccount
\newcommand*\colvec[1]{
        \global\colveccount#1
        \begin{bmatrix}
        \colvecnext
}
\def\colvecnext#1{
        #1
        \global\advance\colveccount-1
        \ifnum\colveccount>0
                \\
                \expandafter\colvecnext
        \else
                \end{bmatrix}
        \fi
}

\newcommand{\shift}{\textsc{shift}}
\newcommand{\reduce}{\textsc{reduce}}

\def\ii#1{\textit{#1}}
\newcommand{\word}[1]{\emph{#1}}

\newcommand{\snli}[3]{{\vspace{0.25em}
{\small \setlength{\parindent}{0.6em} \hangindent=1.2em  \textbf{Premise:} #1\par}\vspace{0.25em}
{\small \setlength{\parindent}{0.6em} \hangindent=1.2em   \textbf{Hypothesis:} #2\par}\vspace{0.25em}
{\small \setlength{\parindent}{0.6em}  \textbf{Label:} #3\par}
}}

\providecommand{\norm}[1]{\lVert#1\rVert}

\noautomath

\title{A Fast Unified Model for Parsing and Sentence Understanding}

\author{
Samuel R.\ Bowman$^{1,2,3,}$\thanks{~\,The first two authors contributed equally.} \\
\texttt{\small sbowman@stanford.edu} \\
\And
Jon Gauthier$^{2,3,4,*}$ \\
\texttt{\small jgauthie@stanford.edu} \\
\And
Abhinav Rastogi$^{3,5}$ \\
\texttt{\small arastogi@stanford.edu} \\
\AND
Raghav Gupta$^{2,3,6}$ \\
\texttt{\small rgupta93@stanford.edu} \\
\And
Christopher D.\ Manning$^{1,2,3,6}$\\
\texttt{\small manning@stanford.edu}\\
\And
Christopher Potts$^{1,6}$\\
\texttt{\small cgpotts@stanford.edu}
\AND\\[-3ex]
{$^{1}$Stanford Linguistics\quad
$^{2}$Stanford NLP Group\quad
$^{3}$Stanford AI Lab}\\
{$^{4}$Stanford Symbolic Systems\quad
$^{5}$Stanford Electrical Engineering\quad
$^{6}$Stanford Computer Science}
}

\date{}

\begin{document}
\maketitle
\begin{abstract}

Tree-structured neural networks exploit valuable syntactic parse information as they interpret the meanings of sentences. However, they suffer from two key technical problems that make them slow and unwieldy for large-scale NLP tasks: they usually operate on parsed sentences and they do not directly support batched computation. We address these issues by introducing the Stack-augmented Parser-Interpreter Neural Network (SPINN), which combines parsing and interpretation within a single tree-sequence hybrid model by integrating tree-structured sentence interpretation into the linear sequential structure of a shift-reduce parser. Our model supports batched computation for a speedup of up to 25$\times$ over other tree-structured models, and its integrated parser can operate on unparsed data with little loss in accuracy. We evaluate it on the Stanford NLI entailment task and show that it significantly outperforms other sentence-encoding models.
\end{abstract}

\section{Introduction}


\begin{figure}[t]

\begin{subfigure}[t]{\columnwidth}
\begin{center}
\scalebox{1}{\includegraphics{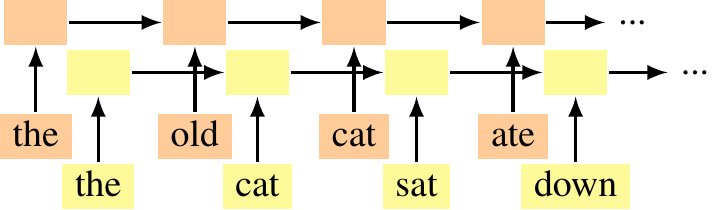}}
\end{center}

\caption{\label{fig:batching:good}A conventional sequence-based RNN for two sentences.}
\end{subfigure}

\begin{subfigure}[t]{\columnwidth}
\begin{center}
\scalebox{1}{
 \includegraphics{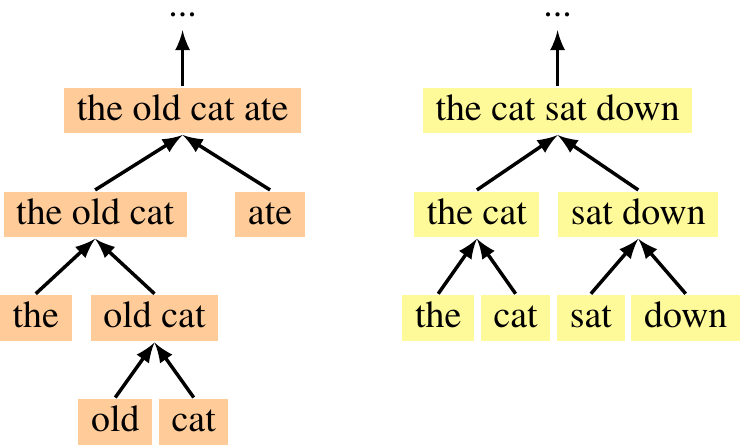}}
\end{center}

\caption{\label{fig:batching:bad}A conventional TreeRNN for two sentences.}
\end{subfigure}

\caption{\label{fig:batching} An illustration of two standard designs for sentence encoders. The TreeRNN, unlike the sequence-based RNN, requires a substantially different connection structure for each sentence, making batched computation impractical.}
\end{figure}


\begin{figure*}[t]
\begin{subfigure}[t]{\textwidth}
\centering
\scalebox{0.8}{\includegraphics{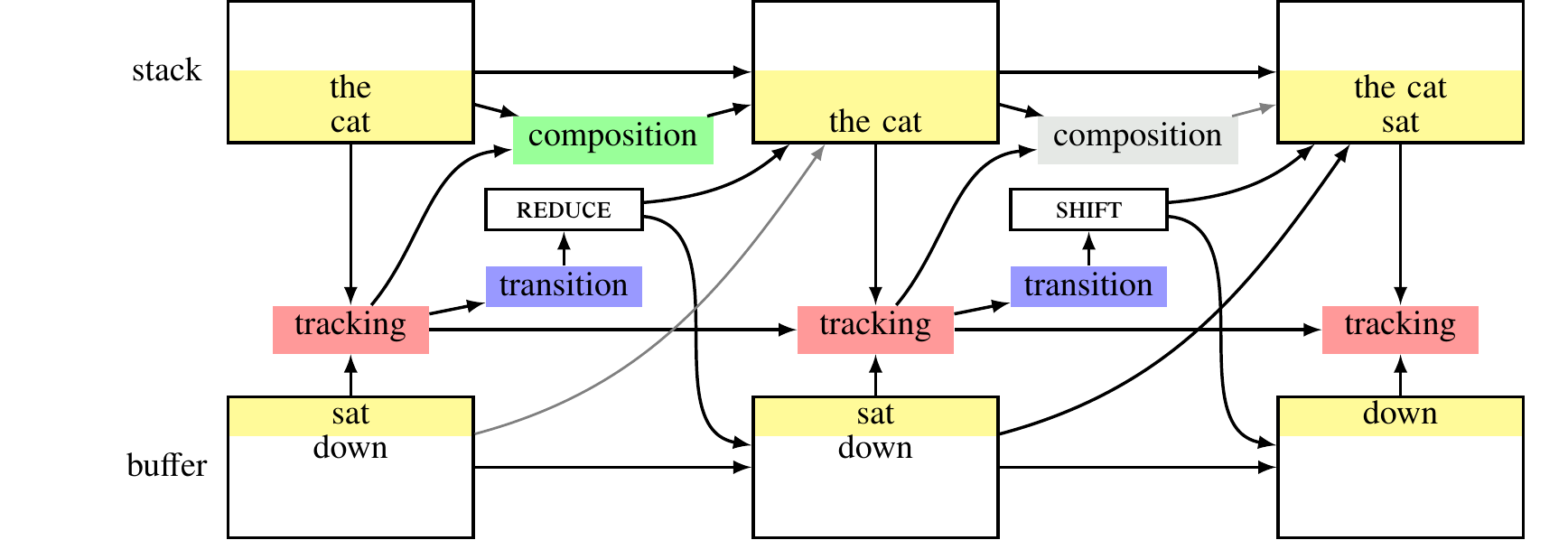}}
  
 \caption{The SPINN model unrolled for two transitions during the processing of the sentence \word{the cat sat down}. `Tracking', `transition', and `composition' are neural network layers. Gray arrows indicate connections which are blocked by a gating function.}\label{fig:model:1d}
  
\end{subfigure}\\\\\\
\begin{subfigure}[t]{\textwidth}
\centering
\scalebox{0.5}{\includegraphics{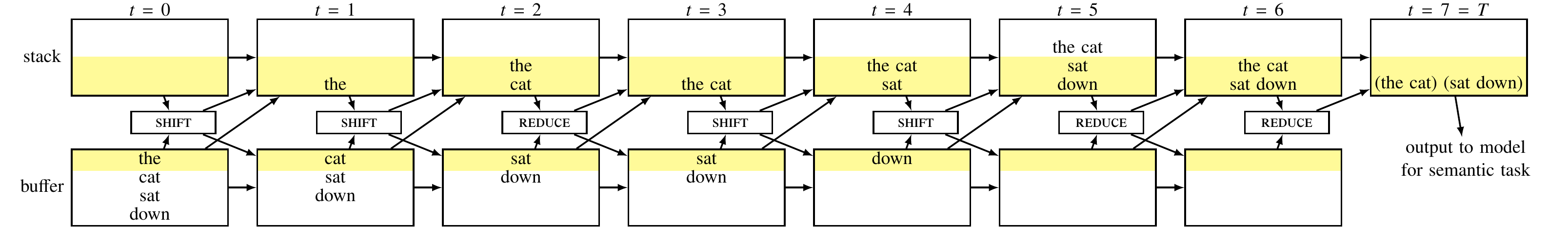}}
  
 \caption{The fully unrolled SPINN for \word{the cat sat down}, with neural network layers omitted for clarity.}\label{fig:model:1b}  
\end{subfigure}
\caption{\label{fig:m1-views}Two views of the Stack-augmented Parser-Interpreter Neural Network (SPINN).}
\end{figure*}

A wide range of current models in NLP are built around a neural network component that produces vector representations of sentence meaning \citep[e.g.,][]{sutskever2014sequence,tai2015improved}. This component, the sentence encoder, is generally formulated as a learned parametric function from a sequence of word vectors to a sentence vector, and this function can take a range of different forms. Common sentence encoders include sequence-based recurrent neural network models (RNNs, see Figure~\ref{fig:batching:good}) with Long Short-Term Memory \citep[LSTM,][]{hochreiter1997long}, which accumulate information over the sentence sequentially; convolutional neural networks \citep{kalchbrenner2014convolutional,DBLP:journals/corr/ZhangZL15}, which accumulate information using filters over short local sequences of words or characters; and tree-structured recursive neural networks \citep[TreeRNNs,][see Figure~\ref{fig:batching:bad}]{goller1996learning,socher2011parsing}, which propagate information up a binary parse tree.

Of these, the TreeRNN appears to be the principled choice, since meaning in natural language sentences is known to be constructed recursively according to a tree structure \citep[][i.a.]{Dowty07DC}. TreeRNNs have shown promise \citep{tai2015improved,li2015tree,bowman2015trees}, but have largely been overlooked in favor of sequence-based RNNs because of their incompatibility with batched computation and their reliance on external parsers.  Batched computation---performing synchronized computation across many examples at once---yields order-of-magnitude improvements in model run time, and is crucial in enabling neural networks to be trained efficiently on large datasets. Because TreeRNNs use a different model structure for each sentence, as in Figure~\ref{fig:batching}, efficient batching is impossible in standard implementations. Partly to address efficiency problems, standard TreeRNN models commonly only operate on sentences that have already been processed by a syntactic parser, which slows and complicates the use of these models at test time for most applications.

This paper introduces a new model to address both these issues: the Stack-augmented Parser-Interpreter Neural Network, or SPINN, shown in Figure~\ref{fig:m1-views}. SPINN executes the computations of a tree-structured model in a linearized sequence, and can incorporate a neural network parser that produces the required parse structure on the fly. This design improves upon the TreeRNN architecture in three ways: At test time, it can simultaneously parse and interpret unparsed sentences, removing the dependence on an external parser at nearly no additional computational cost. Secondly, it supports batched computation for both parsed and unparsed sentences, yielding dramatic speedups over standard TreeRNNs. Finally, it supports a novel tree-sequence hybrid architecture for handling local linear context in sentence interpretation. This model is a basically plausible model of human sentence processing and yields substantial accuracy gains over pure sequence- or tree-based models. 

We evaluate SPINN on the Stanford Natural Language Inference entailment task \citep[SNLI,][]{snli:emnlp2015}, and find that it significantly outperforms other sentence-encoding-based models, even with a relatively simple and underpowered implementation of the built-in parser. We also find that SPINN yields speed increases of up to 25$\times$ over a standard TreeRNN implementation.

\section{Related work}

There is a fairly long history of work on building neural network-based parsers that use the core operations and data structures from transition-based parsing, of which shift-reduce parsing is a variant \citep{henderson2004discriminative,emami2005neural,titov2010latent,chen2014,buys2generative,dyer-EtAl:2015:ACL-IJCNLP,kiperwasser2016easy}. In addition, there has been recent work proposing models designed primarily for generative language modeling tasks that use this architecture as well \citep{zhang2016top,dyer2016rnn}. To our knowledge, SPINN is the first model to use this architecture for the purpose of sentence interpretation, rather than parsing or generation.  

\citet{socher2011parsing,socher2011semi} present versions of the TreeRNN model which are capable of operating over unparsed inputs. However, these methods require an expensive search process at test time. Our model presents a much faster alternative approach.

\section{Our model: SPINN}

\subsection{Background: Shift-reduce parsing}

SPINN is inspired by shift-reduce parsing \citep{aho1972theory}, which builds a tree structure over a sequence (e.g., a natural language sentence) by a single left-to-right scan over its tokens. The formalism is widely used in natural language parsing \citep[e.g.,][]{shieber:1983:ACL,nivre2003efficient}.

A shift-reduce parser accepts a sequence of input tokens $\mathbf x = (x_0, \dots, x_{N-1})$ and consumes transitions $\mathbf a = (a_0, \dots, a_{T-1})$, where each $a_t \in \{\shift, \reduce\}$ specifies one step of the parsing process. In general a parser may also generate these transitions on the fly as it reads the tokens. It proceeds left-to-right through a transition sequence, combining the input tokens $\mathbf x$ incrementally into a tree structure. For any binary-branching tree structure over $N$ words, this requires $T=2N-1$ transitions through a total of $T+1$ states.

The parser uses two auxiliary data structures: a stack $S$ of partially completed subtrees and a buffer $B$ of tokens yet to be parsed. The parser is initialized with the stack empty and the buffer containing the tokens $\mathbf x$ of the sentence in order. Let $\langle S, B \rangle = \langle \emptyset, \mathbf x \rangle$ denote this starting state. It next proceeds through the transition sequence, where each transition $a_t$ selects one of the two following operations. Below, the $\mid$ symbol denotes the \textit{cons} (concatenation) operator. We arbitrarily choose to always \textit{cons} on the left in the notation below.
\vspace*{-1ex} 
\begin{description}
  \item[\shift:] $\langle S, x \mid B \rangle \to \langle x \mid S, B \rangle$. This operation pops an element from the buffer and pushes it on to the top of the stack.
\vspace*{-1ex}
  \item[\reduce:] $\langle x \mid y \mid S, B \rangle \to \langle (x, y) \mid S, B \rangle$. This operation pops the top two elements from the stack, merges them, and pushes the result back on to the stack.
\end{description}

\subsection{Composition and representation}

SPINN is based on a shift-reduce parser, but it is designed to produce a vector representation of a sentence as its output, rather than a tree as in standard shift-reduce parsing. It modifies the shift-reduce formalism by using fixed length vectors to represent each entry in the stack and the buffer. Correspondingly, its \reduce\ operation combines two vector representations from the stack into another vector using a neural network function.

\paragraph{The composition function}
When a \reduce\ operation is performed, the vector representations of two tree nodes are popped off of the stack and fed into a {\it composition function}, which is a neural network function that produces a representation for a new tree node that is the parent of the two popped nodes. This new node is pushed on to the stack.

The TreeLSTM composition function \citep{tai2015improved} generalizes the LSTM neural network layer to tree- rather than sequence-based inputs, and it shares with the LSTM the idea of representing intermediate states as a pair of an active state representation $\vec{h}$ and a memory representation $\vec{c}$. Our version is formulated as:
\begin{gather}
\colvec{5}
    {\vec{i}}
    {\vec{f}_l}
    {\vec{f}_r}
    {\vec{o}}
    {\vec{g}}
= \colvec{5}
     {\sigma\vphantom{\vec{i}}}
     {\sigma\vphantom{\vec{f}_l}}
     {\sigma\vphantom{\vec{f}_r}}
     {\sigma\vphantom{\vec{o}}}
    {\text{tanh}\vphantom{\vec{g}}}
\left(
W_{\text{comp}}
\colvec{3}
    {\vec{h}_s^1}
    {\vec{h}_s^2}
    {\vec{e}}
+ \vec{b}_{\text{comp}}
\right) \label{eqn:lstm1}
\\
\vec{c} = \vec{f}_l \odot \vec{c}_s^{\,2} + \vec{f}_r \odot \vec{c}_s^{\,1} + \vec{i} \odot \vec{g}
\\
\vec{h} = \vec{o} \odot \tanh(\vec{c})
\end{gather}
where $\sigma$ is the sigmoid activation function, $\odot$ is the elementwise product, the pairs $\langle\vec{h}^1_s, \vec{c}^{\,1}_s\rangle$ and $\langle\vec{h}^2_s, \vec{c}^{\,2}_s\rangle$ are the two input tree nodes popped off the stack, and $\vec{e}$ is an optional vector-valued input argument which is either empty or comes from an external source like the tracking LSTM (see Section~\ref{sec:tracking}). The result of this function, the pair $\langle\vec{h}, \vec{c}\rangle$, is placed back on the stack. Each vector-valued variable listed is of dimension $D$ except $\vec{e}$, of the independent dimension $D_{\text{tracking}}$.

\paragraph{The stack and buffer}

The stack and the buffer are arrays of $N$ elements each (for sentences of up to $N$ words), with the two $D$-dimensional vectors $\vec{h}$ and $\vec{c}$ in each element.

\paragraph{Word representations}

We use word representations based on the 300D vectors provided with GloVe \citep{pennington2014glove}. We do not update these representations during training. Instead, we use a learned linear transformation to map each input word vector $\vec{x}_{\text{GloVe}}$ into a vector pair $\langle \vec{h}, \vec{c}\rangle$ that is stored in the buffer:
\begin{equation}
\colvec{2}
    {\vec{h}}
    {\vec{c}}
= W_{\text{wd}} \vec{x}_{\text{GloVe}} + \vec{b}_{\text{wd}}
\end{equation}

\subsection{The tracking LSTM}\label{sec:tracking}

In addition to the stack, buffer, and composition function, our full model includes an additional component: the tracking LSTM. This is a simple sequence-based LSTM RNN that operates in tandem with the model, taking inputs from the buffer and stack at each step. It is meant to maintain a low-resolution summary of the portion of the sentence that has been processed so far, which is used for two purposes: it supplies feature representations to the transition classifier, which allows the model to stand alone as a parser, and it additionally supplies a secondary input $\vec{e}$ to the composition function---see (\ref{eqn:lstm1})---allowing context information to enter the construction of sentence meaning and forming what is effectively a tree-sequence hybrid model.

The tracking LSTM's inputs (yellow in Figure~\ref{fig:m1-views}) are the top element of the buffer $\vec{h}_b^1$ (which would be moved in a \shift\ operation) and the top two elements of the stack $\vec{h}_s^1$ and $\vec{h}_s^2$ (which would be composed in a \reduce\ operation).

\paragraph{Why a tree-sequence hybrid?}

Lexical ambiguity is ubiquitous in natural language. Most words have multiple senses or meanings, and it is generally necessary to use the context in which a word occurs to determine which of its senses or meanings is meant in a given sentence. Even though TreeRNNs are more effective at composing meanings in principle, this ambiguity can give simpler sequence-based sentence-encoding models an advantage: when a sequence-based model first processes a word, it has direct access to a state vector that summarizes the left context of that word, which acts as a cue for disambiguation. In contrast, when a standard tree-structured model first processes a word, it only has access to the constituent that the word is merging with, which is often just a single additional word. Feeding a context representation from the tracking LSTM into the composition function is a simple and efficient way to mitigate this disadvantage of tree-structured models. Using left linear context to disambiguate is also a plausible model of human interpretation.

It would be straightforward to augment SPINN to support the use of some amount of right-side context as well, but this would add complexity to the model that we think is largely unnecessary: humans are very effective at understanding the beginnings of sentences before having seen or heard the ends, suggesting that it is possible to get by without the unavailable right-side context.

\subsection{Parsing: Predicting transitions}

For SPINN to operate on unparsed inputs, it needs to produce its own transition sequence $\mathbf a$ rather than relying on an external parser to supply it as part of the input. To do this, the model predicts $a_t$ at each step using a simple two-way softmax classifier whose input is the state of the tracking LSTM:
\begin{equation}
\vec{p}_{\text{a}} = \text{softmax}(W_{\text{trans}}\vec{h}_{\text{tracking}} + \vec{b}_{\text{trans}})
\end{equation}
The above model is nearly the simplest viable implementation of a transition decision function. In contrast, the decision functions in state-of-the-art transition-based parsers tend to use significantly richer feature sets as inputs, including features containing information about several upcoming words on the buffer. The value $\vec{h}_{\text{tracking}}$ is a function of only the very top of the buffer and the top two stack elements at each timestep.

At test time, the model uses whichever transition (i.e., \shift\ or \reduce) is assigned a higher (unnormalized) probability. The prediction function is trained to mimic the decisions of an external parser. These decisions are used as inputs to the model during training. For SNLI, we use the binary Stanford PCFG Parser parses that are included with the corpus. We did not find scheduled sampling \citep{bengio2015scheduled}---having the model use its own transition decisions sometimes at training time---to help.

\subsection{Implementation issues}

\paragraph{Representing the stack efficiently}

A na\"ive implementation of SPINN needs to handle a size $O(N)$ stack at each timestep, any element of which may be involved in later computations. A na\"ive backpropagation implementation would then require storing each of the $O(N)$ stacks for a backward pass, leading to a per-example space requirement of $O(NTD)$ floats. This requirement is prohibitively large for significant batch sizes or sentence lengths $N$. Such a na\"ive implementation would also require copying a largely unchanged stack at each timestep, since each \shift\ or \reduce\ operation writes only one new representation to the top of the stack.

We propose a space-efficient stack representation inspired by the zipper technique \citep{huet1997zipper} that we call \textit{thin stack}. For each input sentence, we represent the stack with a single $T \times D$ matrix $S$. Each row $S[t]$ (for $0<t\le T$) represents the top of the actual stack at timestep $t$. At each timestep we can \shift\ a new element onto the stack, or \reduce\ the top two elements of the stack into a single element. To shift an element from the buffer to the top of the stack at timestep $t$, we simply write it into the location $S[t]$. In order to perform the \reduce\ operation, we need to retrieve the top two elements of the actual stack. We maintain a queue $Q$ of pointers into $S$ which contains the row indices of $S$ which are still present in the actual stack. The top two elements of the stack can be found by using the final two pointers in the queue $Q$. These retrieved elements are used to perform the \reduce\ operation, which modifies $Q$ to mark that some rows of $S$ have now been replaced in the actual stack. Algorithm~\ref{alg:thin-stack} describes the full mechanics of a stack feedforward in this compressed representation. It operates on the single $T \times D$ matrix $S$ and a backpointer queue $Q$. Table~\ref{tbl:thin-stack} shows an example run.

\begin{algorithm}[t]
\caption{The thin stack algorithm}
\label{alg:thin-stack}
\begin{algorithmic}[1]
  \Function{Step}{bufferTop, $a$, $t$, $S$, $Q$}
    \If{$a$ = \shift}
      \State $S$[$t$] := bufferTop
    \ElsIf{$a$ = \reduce}
      \State right := $S$[$Q$.pop()]
      \State left := $S$[$Q$.pop()]
      \State $S$[$t$] := \Call{Compose}{left, right}
    \EndIf
    \State $Q$.push($t$)
  \EndFunction
\end{algorithmic}
\end{algorithm}

This stack representation requires substantially less space. It stores each element involved in the feedforward computation exactly once, meaning that this representation can still support efficient backpropagation. Furthermore, all of the updates to $S$ and $Q$ can be performed batched and in-place on a GPU, yielding substantial speed gains over both a more na\"ive SPINN implementation and a standard TreeRNN implementation. We describe speed results in Section~\ref{sec:speed}.

\paragraph{Preparing the data} At training time, SPINN requires both a transition sequence $\mathbf a$  and a token sequence $\mathbf x$ as its inputs for each sentence. The token sequence is simply the words in the sentence in order. $\mathbf a$ can be obtained from any constituency parse for the sentence by first converting that parse into an unlabeled binary parse, then linearizing it (with the usual in-order traversal), then taking each word token as a \shift\ transition and each `)' as a \reduce\ transition, as here:

\begin{table}[t]
\centering
\begin{tabular}{c @{\hspace*{1.5em}}c @{\hspace*{1.5em}} lc}
  \toprule
  $t$ & $S$[$t$] & \multicolumn{1}{c}{$Q_t$} & $a_t$ \\
  \midrule
  0 &   & \underline{\hphantom{0} \hphantom{0}} & \shift \\
  1 &  \word{Spot} & \hphantom{0} \underline{\hphantom{0} 1} & \shift \\
  2 &  \word{sat} & \hphantom{0} \hphantom{0} \underline{1 2} & \shift \\
  3  &   \word{down} & \hphantom{0} \hphantom{0} 1 \underline{2 3} & \reduce\\
  4  &   (\word{sat down}) & \hphantom{0} \hphantom{0} \underline{1 4} & \reduce\\
  5  & (\word{Spot} (\word{sat down})) & \hphantom{0} \underline{\hphantom{0} 5} &        \\
  \bottomrule
\end{tabular}
\caption{The thin-stack algorithm operating on the input sequence $\mathbf x = \text{(\word{Spot}, \word{sat}, \word{down})}$ and the transition sequence shown in the rightmost column. $S$[$t$] shows the top of the stack at each step $t$. The last two elements of $Q$ (underlined) specify which rows $t$ would be involved in a \reduce\ operation at the next step.}
\label{tbl:thin-stack}
\end{table}

\vspace{0.5em}
{\noindent\small
{\bf Unlabeled binary parse:} ( ( the cat ) ( sat down ) )\\
{$\mathbf x$}: \word{the}, \word{cat}, \word{sat}, \word{down}\\
{$\mathbf a$}: \shift, \shift, \reduce, \shift, \shift, \reduce, \reduce
}

\paragraph{Handling variable sentence lengths} For any sentence model to be trained with batched computation, it is necessary to pad or crop sentences to a fixed length. We fix this length at $N = 25$ words, longer than about 98\% of sentences in SNLI\@. Transition sequences $\mathbf a$ are cropped at the left or padded at the left with \shift s. Token sequences $\mathbf x$ are then cropped or padded with empty tokens at the left to match the number of \shift s added or removed from $\mathbf a$, and can then be padded with empty tokens at the right to meet the desired length $N$.

\subsection{TreeRNN-equivalence}

Without the addition of the tracking LSTM, SPINN (in particular the SPINN-PI-NT variant, for \textit{parsed input, no tracking}) is precisely equivalent to a conventional tree-structured neural network model in the function that it computes, and therefore it also has the same learning dynamics. In both, the representation of each sentence consists of the representations of the words combined recursively using a TreeRNN composition function (in our case, the TreeLSTM function). SPINN, however, is dramatically faster, and supports both integrated parsing and a novel approach to context through the tracking LSTM.

\subsection{Inference speed}
\label{sec:speed}

In this section, we compare the test-time speed of our SPINN-PI-NT with an equivalent TreeRNN implemented in the conventional fashion and with a standard RNN sequence model. While the full models evaluated below are implemented and trained using Theano \citep{theano}, which is reasonably efficient but not perfect for our model, we wish to compare well-optimized implementations of all three models. To do this, we reimplement the feedforward\footnote{We chose to reimplement and evaluate only the feedforward/inference pass, as inference speed is the relevant performance metric for most practical applications.} of SPINN-PI-NT and an LSTM RNN baseline in C++/CUDA, and compare that implementation with a CPU-based C++/Eigen TreeRNN implementation from \citet{irsoy2014deep}, which we modified to perform exactly the same computations as SPINN-PI-NT.\footnote{The original code for Irsoy \& Cardie's model is available at \url{https://github.com/oir/deep-recursive}. Our optimized C++/CUDA models and the Theano source code for the full SPINN are available at \url{https://github.com/stanfordnlp/spinn}.} TreeRNNs like this can only operate on a single example at a time and are thus poorly suited for GPU computation.

\begin{figure}
\centering
\resizebox{1\linewidth}{!}{
\includegraphics{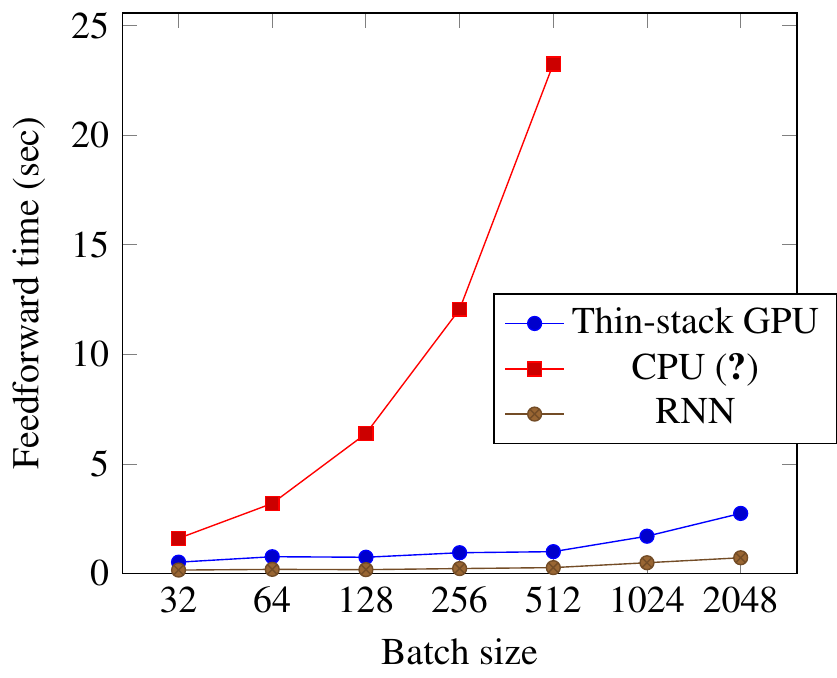}
}
\caption{Feedforward speed comparison.}
\label{fig:speed}
\end{figure}

Each model is restricted to run on sentences of 30 tokens or fewer. We fix the model dimension $D$ and the word embedding dimension at 300. We run the CPU performance test on a 2.20 GHz 16-core Intel Xeon E5-2660 processor with hyperthreading enabled. We test our thin-stack implementation and the RNN model on an NVIDIA Titan X GPU.

Figure~\ref{fig:speed} compares the sentence encoding speed of the three models on random input data. We observe a substantial difference in runtime between the CPU and thin-stack implementations that increases with batch size. With a large but practical batch size of 512, the largest on which we tested the TreeRNN, our model is about 25$\times$ faster than the standard CPU implementation, and about 4$\times$ slower than the RNN baseline.

Though this experiment only covers SPINN-PI-NT, the results should be similar for the full SPINN model: most of the computation involved in running SPINN is involved in populating the buffer, applying the composition function, and manipulating the buffer and the stack, with the low-dimensional tracking and parsing components adding only a small additional load.

\begin{table*}[t]\small
\begin{center}
\begin{tabular}{lrlrrrr}
\toprule
Param.     & Range & Strategy        & RNN       & SP.-PI-NT   & SP.-PI  & SP. \\
\midrule 
Initial LR & $2\times 10^{-4}$--$2\times 10^{-2}$ & \textsc{log} & $5\times 10^{-3}$  & $3\times 10^{-4}$ & $7\times 10^{-3}$  & $2\times 10^{-3}$\\
L2 regularization $\lambda$ & $8\times 10^{-7}$--$3\times 10^{-5}$   & \textsc{log} & $4\times 10^{-6}$  & $3\times 10^{-6}$   & $2\times 10^{-5}$  & $3\times 10^{-5}$\\
Transition cost $\alpha$  & 0.5--4.0 & \textsc{lin} & --- & --- & ---  & 3.9    \\
Embedding transformation dropout & 80--95\% & \textsc{lin} & --- & 83\% & 92\%  & 86\%\\
Classifier MLP dropout & 80--95\% & \textsc{lin} & 94\%  & 94\%   & 93\%  & 94\%\\
Tracking LSTM size $D_\text{tracking}$ & 24--128 & \textsc{log} & --- & --- & 61  & 79\\
Classifier MLP layers & 1--3 & \textsc{lin} & 2 & 2 & 2 & 1\\
\bottomrule
\end{tabular}
\end{center}
\caption{
\label{tab:hyperparams}
Hyperparameter ranges and values. \ii{Range} shows the hyperparameter ranges explored during random search. \ii{Strategy} indicates whether sampling from the range was uniform, or log-uniform. Dropout parameters are expressed as keep rates rather than drop rates.
}
\end{table*}

\section{NLI Experiments}

We evaluate SPINN on the task of natural language inference \citep[NLI, a.k.a.\ recognizing textual entailment, or RTE;][]{dagan2006pascal}. NLI is a sentence pair classification task, in which a model reads two sentences (a premise and a hypothesis), and outputs a judgment of {\it entailment}, {\it contradiction}, or {\it neutral}, reflecting the relationship between the meanings of the two sentences. Below is an example sentence pair and judgment from the SNLI corpus which we use in our experiments:

\snli{Girl in a red coat, blue head wrap and jeans is making a snow angel.}
{A girl outside plays in the snow.}
{entailment}

SNLI is a corpus of 570k human-labeled pairs of scene descriptions like this one. We use the standard train--test split and ignore unlabeled examples, which leaves about 549k examples for training, 9,842 for development, and 9,824 for testing. SNLI labels are roughly balanced, with the most frequent label, {\it entailment}, making up 34.2\% of the test set.

Although NLI is framed as a simple three-way classification task, it is nonetheless an effective way of evaluating the ability of a model to extract broadly informative representations of sentence meaning. In order for a model to perform reliably well on NLI, it must be able to represent and reason with the core phenomena of natural language semantics, including quantification, coreference, scope, and several types of ambiguity.

\subsection{Applying SPINN to SNLI}

\paragraph{Creating a sentence-pair classifier} \label{sec:classifier}

To classify an SNLI sentence pair, we run two copies of SPINN with shared parameters: one on the premise sentence and another on the hypothesis sentence. We then use their outputs (the $\vec{h}$ states at the top of each stack at time $t=T$) to construct a feature vector $\vec{x}_{\text{classifier}}$ for the pair. This feature vector consists of the concatenation of these two sentence vectors, their difference, and their elementwise product \citep[following][]{mou2015recognizing}:
\begin{equation}
\vec{x}_{\text{classifier}} =
\colvec{4}
    {\vec{h}_{\text{premise}}}
    {\vec{h}_{\text{hypothesis}}}
    {\vec{h}_{\text{premise}} - \vec{h}_{\text{hypothesis}}}
    {\vec{h}_{\text{premise}} \odot \vec{h}_{\text{hypothesis}}}
\end{equation}
This feature vector is then passed to a series of 1024D ReLU neural network layers (i.e., an MLP; the number of layers is tuned as a hyperparameter), then passed into a linear transformation, and then finally passed to a softmax layer, which yields a distribution over the three labels.

\paragraph{The objective function} Our objective combines a cross-entropy objective $\mathcal{L}_{\text{s}}$ for the SNLI classification task, cross-entropy objectives $\mathcal{L}^t_{\text{p}}$ and $\mathcal{L}^t_{\text{h}}$ for the parsing decision for each of the two sentences at each step $t$, and an L2 regularization term on the trained parameters. The terms are weighted using the tuned hyperparameters $\alpha$ and $\lambda$:
\begin{equation}
\begin{split}
\mathcal{L}_{\text{m}} = &\mathcal{L}_{\text{s}} + \alpha \sum_{t=0}^{T-1} (\mathcal{L}^t_{\text{p}} + \mathcal{L}^t_{\text{h}}) + \lambda \norm{\theta}^2_2
\end{split}
\end{equation}

\paragraph{Initialization, optimization, and tuning}

We initialize the model parameters using the nonparametric strategy of \citet{DBLP:journals/corr/HeZR015}, with the exception of the softmax classifier parameters, which we initialize using random uniform samples from $[-0.005, 0.005]$.
 
We use minibatch SGD with the RMSProp optimizer \citep{tieleman2012lecture} and a tuned starting learning rate that decays by a factor of 0.75 every 10k steps. We apply both dropout \citep{srivastava2014dropout} and batch normalization \citep{2015SIoffeCSzegedy} to the output of the word embedding projection layer and to the feature vectors that serve as the inputs and outputs to the MLP that precedes the final entailment classifier.

We train each model for 250k steps in each run, using a batch size of 32. We track each model's performance on the development set during training and save parameters when this performance reaches a new peak. We use early stopping, evaluating on the test set using the parameters that perform best on the development set.

We use random search to tune the hyperparameters of each model, setting the ranges for search for each hyperparameter heuristically (and validating the reasonableness of the ranges on the development set), and then launching eight copies of each experiment each with newly sampled hyperparameters from those ranges. Table~\ref{tab:hyperparams} shows the hyperparameters used in the best run of each model.

\begin{table*}[t]
  \centering\small
  \begin{tabular}{lrrrr}
    \toprule
Model                   & Params.    & Trans. acc. (\%)  &   Train acc. (\%)  &   Test acc. (\%) \\
\midrule
\multicolumn{5}{c}{\textbf{Previous non-NN results}}\\
Lexicalized classifier \citep{snli:emnlp2015}
                        & ---                & ---                    &   99.7   &   78.2      \\
\midrule
\multicolumn{5}{c}{\textbf{Previous sentence encoder-based NN results}}\\
100D LSTM encoders \citep{snli:emnlp2015}
                        & 221k               & ---               &   84.8   &   77.6      \\
1024D pretrained GRU encoders \citep{DBLP:journals/corr/VendrovKFU15}
                        & 15m                & ---              &   98.8   &   81.4       \\
300D Tree-based CNN encoders \citep{mou2015recognizing}
                        & 3.5m                & ---             &   83.4   &   82.1       \\
\midrule
\multicolumn{5}{c}{\textbf{Our results}}\\
300D LSTM RNN encoders         & 3.0m                  & ---                &   83.9      &   80.6       \\
300D SPINN-PI-NT (\ii{parsed input, no tracking}) encoders
                        & 3.4m                  & ---                &   84.4      &   80.9       \\
300D SPINN-PI (\ii{parsed input}) encoders
                        & 3.7m                  & ---                &   89.2      &   \textbf{83.2}       \\
300D SPINN (unparsed input) encoders
                        & 2.7m                  & 92.4           &   87.2    &   82.6      \\
    \bottomrule
  \end{tabular}
\caption{\protect\label{tab:results}Results on SNLI 3-way inference classification. Params.\ is the approximate number of trained parameters (excluding word embeddings for all models). Trans.\ acc.\ is the model's accuracy in predicting parsing transitions at test time. Train and test are SNLI classification accuracy.}
\end{table*}

\subsection{Models evaluated}

We evaluate four models. The four all use the sentence-pair classifier architecture described in Section~\ref{sec:classifier}, and differ only in the function computing the sentence encodings. First, a single-layer LSTM RNN \citep[similar to that of][]{snli:emnlp2015} serves as a baseline encoder. Next, the minimal SPINN-PI-NT model (equivalent to a TreeLSTM) introduces the SPINN model design. SPINN-PI adds the tracking LSTM to that design. Finally, the full SPINN adds the integrated parser.

We compare our models against several baselines, including the strongest published non-neural network-based result from \citet{snli:emnlp2015} and previous neural network models built around several types of sentence encoders.

\subsection{Results}

Table~\ref{tab:results} shows our results on SNLI. For the full SPINN, we also report a measure of agreement between this model's parses and the parses included with SNLI, calculated as classification accuracy over transitions averaged across timesteps.

We find that the bare SPINN-PI-NT model performs little better than the RNN baseline, but that SPINN-PI with the added tracking LSTM performs well. The success of SPINN-PI, which is the hybrid tree-sequence model, suggests that the tree- and sequence-based encoding methods are at least partially complementary, with the sequence model presumably providing useful local word disambiguation. The full SPINN model with its relatively weak internal parser performs slightly less well, but nonetheless robustly exceeds the performance of the RNN baseline.

Both SPINN-PI and the full SPINN significantly outperform all previous sentence-encoding models. Most notably, these models outperform the tree-based CNN of \citet{mou2015recognizing}, which also uses tree-structured composition for local feature extraction, but uses simpler pooling techniques to build sentence features in the interest of efficiency. Our results show that a model that uses tree-structured composition fully (SPINN) outperforms one which uses it only partially (tree-based CNN), which in turn outperforms one which does not use it at all (RNN).

The full SPINN performed moderately well at reproducing the Stanford Parser's parses of the SNLI data at a transition-by-transition level, with 92.4\% accuracy at test time.\footnote
  {Note that this is scoring the model against automatic parses, not a human-judged gold standard.}
However, its transition prediction errors are fairly evenly distributed across sentences, and most sentences were assigned partially invalid transition sequences that either left a few words out of the final representation or incorporated a few padding tokens into the final representation.

\subsection{Discussion}

The use of tree structure improves the performance of sentence-encoding models for SNLI. We suspect that this improvement is largely due to the more efficient learning of accurate generalizations overall, and not to any particular few phenomena. However, some patterns are identifiable in the results. 

While all four models under study have trouble with negation, the tree-structured SPINN models do quite substantially better on these pairs. This is likely due to the fact that parse trees make the scope of any instance of negation (the portion of the sentence's content that is negated) relatively easy to identify and separate from the rest of the sentence. For test set sentence pairs like the one below where negation (\word{not} or \word{n't}) does not appear in the premise but does appear in the hypothesis, the RNN shows $67\%$ accuracy, while all three tree-structured models exceed $73\%$. Only the RNN got the below example wrong:

\snli
{The rhythmic gymnast completes her floor exercise at the competition.}
{The gymnast cannot finish her exercise.}
{contradiction}
\noindent Note that the presence of negation in the hypothesis is correlated with a label of \textit{contradiction} in SNLI, but not as strongly as one might intuit---only $45\%$ of these examples in the test set are labeled as contradictions.

In addition, it seems that tree-structured models, and especially the tree-sequence hybrid models, are more effective than RNNs at extracting informative representations of long sentences. The RNN model falls off in test accuracy more quickly with increasing sentence length than SPINN-PI-NT, which in turn falls of substantially faster than the two hybrid models, repeating a pattern seen more dramatically on artificial data in \citet{bowman2015trees}. On pairs with premises of 20 or more words, the RNN's 76.7\% accuracy, while SPINN-PI reaches 80.2\%. All three SPINN models labeled the following example correctly, while the RNN did not:

\snli
{A man wearing glasses and a ragged costume is playing a Jaguar electric guitar and singing with the accompaniment of a drummer.}
{A man with glasses and a disheveled outfit is playing a guitar and singing along with a drummer.}
{entailment}

We suspect that the hybrid nature of the full SPINN model is also responsible for its surprising ability to perform better than an RNN baseline even when its internal parser is relatively ineffective at producing correct full-sentence parses. It may act somewhat like the tree-based CNN, only with access to larger trees: using tree structure to build up local phrase meanings, and then using the tracking LSTM, at least in part, to combine those meanings.

Finally, as is likely inevitable for models evaluated on SNLI, all four models under study did several percent worse on test examples whose ground truth label is \textit{neutral} than on examples of the other two classes. \textit{Entailment}--\textit{neutral} and \textit{neutral}--\textit{contradiction} confusions appear to be much harder to avoid than \textit{entailment}--\textit{contradiction} confusions, where relatively superficial cues might be more readily useful.

\section{Conclusions and future work}

We introduce a model architecture (SPINN-PI-NT) that is equivalent to a TreeLSTM, but an order of magnitude faster at test time. We expand that architecture into a tree-sequence hybrid model (SPINN-PI), and show that this yields significant gains on the SNLI entailment task. Finally, we show that it is possible to exploit the strengths of this model without the need for an external parser by integrating a fast parser into the model (as in the full SPINN), and that the lack of external parse information yields little loss in accuracy.

Because this paper aims to introduce a general purpose model for sentence encoding, we do not pursue the use of soft attention \citep{bahdanau2014neural,rocktaschel2015reasoning}, despite its demonstrated effectiveness on the SNLI task.\footnote{Attention-based models like \citet{rocktaschel2015reasoning}, \citet{wang2015learning}, and the unpublished \citet{cheng2016long} have shown accuracies as high as 86.3\% on SNLI, but are more narrowly engineered to suit the task and do not yield sentence encodings.} However, we expect that it should be possible to productively combine our model with soft attention to reach state-of-the-art performance. 

Our tracking LSTM uses only simple, quick-to-compute features drawn from the head of the buffer and the head of the stack. It is plausible that giving the tracking LSTM access to more information from the buffer and stack at each step would allow it to better represent the context at each tree node, yielding both better parsing and better sentence encoding. One promising way to pursue this goal would be to encode the full contents of the stack and buffer at each time step following the method used by \citet{dyer-EtAl:2015:ACL-IJCNLP}.

For a more ambitious goal, we expect that it should be possible to implement a variant of SPINN on top of a modified stack data structure with differentiable \textsc{push} and \textsc{pop} operations \citep[as in][]{grefenstette2015learning,joulin2015inferring}. This would make it possible for the model to learn to parse using guidance from the semantic representation objective, which currently is blocked from influencing the key parsing parameters by our use of hard \shift/\reduce\ decisions. This change would allow the model to learn to produce parses that are, in aggregate, better suited to supporting semantic interpretation than those supplied in the training data.

\subsubsection*{Acknowledgments}

We acknowledge financial support from a Google Faculty Research Award, the Stanford Data Science Initiative, and the National Science Foundation under grant nos.~BCS 1456077 and IIS 1514268. Some of the Tesla K40s used for this research were donated by the NVIDIA Corporation. We also thank Kelvin Guu, Noah Goodman, and many others in the Stanford NLP group for helpful comments.

\bibliographystyle{acl_natbib}
\bibliography{MLSemantics}

\begin{thebibliography}{}
\expandafter\ifx\csname natexlab\endcsname\relax\def\natexlab#1{#1}\fi

\bibitem[{Aho and Ullman(1972)}]{aho1972theory}
Alfred~V. Aho and Jeffrey~D. Ullman. 1972.
\newblock {\em The theory of parsing, translation, and compiling\/}.
\newblock Prentice-Hall, Inc.

\bibitem[{Bahdanau et~al.(2015)Bahdanau, Cho, and Bengio}]{bahdanau2014neural}
Dzmitry Bahdanau, Kyunghyun Cho, and Yoshua Bengio. 2015.
\newblock Neural machine translation by jointly learning to align and
  translate.
\newblock In {\em Proc. {ICLR}\/}.

\bibitem[{Bengio et~al.(2015)Bengio, Vinyals, Jaitly, and
  Shazeer}]{bengio2015scheduled}
Samy Bengio, Oriol Vinyals, Navdeep Jaitly, and Noam Shazeer. 2015.
\newblock Scheduled sampling for sequence prediction with recurrent neural
  networks.
\newblock In {\em Proc. NIPS\/}.

\bibitem[{Bowman et~al.(2015{\natexlab{a}})Bowman, Angeli, Potts, and
  Manning}]{snli:emnlp2015}
Samuel~R. Bowman, Gabor Angeli, Christopher Potts, and Christopher~D. Manning.
  2015{\natexlab{a}}.
\newblock A large annotated corpus for learning natural language inference.
\newblock In {\em Proc. EMNLP\/}.

\bibitem[{Bowman et~al.(2015{\natexlab{b}})Bowman, Manning, and
  Potts}]{bowman2015trees}
Samuel~R. Bowman, Christopher~D. Manning, and Christopher Potts.
  2015{\natexlab{b}}.
\newblock Tree-structured composition in neural networks without
  tree-structured architectures.
\newblock In {\em Proc. 2015 NIPS Workshop on Cognitive Computation:
  Integrating Neural and Symbolic Approaches\/}.

\bibitem[{Buys and Blunsom(2015)}]{buys2generative}
Jan Buys and Phil Blunsom. 2015.
\newblock Generative incremental dependency parsing with neural networks.
\newblock In {\em Proc. ACL\/}.

\bibitem[{Chen and Manning(2014)}]{chen2014}
Danqi Chen and Christopher~D. Manning. 2014.
\newblock A fast and accurate dependency parser using neural networks.
\newblock In {\em Proc. EMNLP\/}.

\bibitem[{Cheng et~al.(2016)Cheng, Dong, and Lapata}]{cheng2016long}
Jianpeng Cheng, Li~Dong, and Mirella Lapata. 2016.
\newblock Long short-term memory-networks for machine reading.
\newblock {arXiv}:1601.06733.

\bibitem[{Dagan et~al.(2006)Dagan, Glickman, and Magnini}]{dagan2006pascal}
Ido Dagan, Oren Glickman, and Bernardo Magnini. 2006.
\newblock The {PASCAL} recognising textual entailment challenge.
\newblock In {\em Machine learning challenges. Evaluating predictive
  uncertainty, visual object classification, and recognising tectual
  entailment\/}, Springer.

\bibitem[{Dowty(2007)}]{Dowty07DC}
David Dowty. 2007.
\newblock Compositionality as an empirical problem.
\newblock In {\em Direct Compositionality\/}, Oxford Univ. Press.

\bibitem[{Dyer et~al.(2015)Dyer, Ballesteros, Ling, Matthews, and
  Smith}]{dyer-EtAl:2015:ACL-IJCNLP}
Chris Dyer, Miguel Ballesteros, Wang Ling, Austin Matthews, and Noah~A. Smith.
  2015.
\newblock Transition-based dependency parsing with stack long short-term
  memory.
\newblock In {\em Proc. ACL\/}.

\bibitem[{Dyer et~al.(2016)Dyer, Kuncoro, Ballesteros, and Smith}]{dyer2016rnn}
Chris Dyer, Adhiguna Kuncoro, Miguel Ballesteros, and Noah~A. Smith. 2016.
\newblock Recurrent neural network grammars.
\newblock In {\em Proc. NAACL\/}.

\bibitem[{Emami and Jelinek(2005)}]{emami2005neural}
Ahmad Emami and Frederick Jelinek. 2005.
\newblock A neural syntactic language model.
\newblock {\em Machine learning\/} 60(1--3).

\bibitem[{Goller and K{\"u}chler(1996)}]{goller1996learning}
Christoph Goller and Andreas K{\"u}chler. 1996.
\newblock Learning task-dependent distributed representations by
  backpropagation through structure.
\newblock In {\em Proc. {IEEE} International Conference on Neural Networks\/}.

\bibitem[{Grefenstette et~al.(2015)Grefenstette, Hermann, Suleyman, and
  Blunsom}]{grefenstette2015learning}
Edward Grefenstette, Karl~Moritz Hermann, Mustafa Suleyman, and Phil Blunsom.
  2015.
\newblock Learning to transduce with unbounded memory.
\newblock In {\em Proc. NIPS\/}.

\bibitem[{He et~al.(2015)He, Zhang, Ren, and Sun}]{DBLP:journals/corr/HeZR015}
Kaiming He, Xiangyu Zhang, Shaoqing Ren, and Jian Sun. 2015.
\newblock Delving deep into rectifiers: Surpassing human-level performance on
  {ImageNet} classification.
\newblock In {\em Proc. ICCV\/}.

\bibitem[{Henderson(2004)}]{henderson2004discriminative}
James Henderson. 2004.
\newblock Discriminative training of a neural network statistical parser.
\newblock In {\em Proc. ACL\/}.

\bibitem[{Hochreiter and Schmidhuber(1997)}]{hochreiter1997long}
Sepp Hochreiter and J{\"u}rgen Schmidhuber. 1997.
\newblock Long short-term memory.
\newblock {\em Neural computation\/} 9(8).

\bibitem[{Huet(1997)}]{huet1997zipper}
G{\'e}rard Huet. 1997.
\newblock The zipper.
\newblock {\em Journal of functional programming\/} 7(5).

\bibitem[{Ioffe and Szegedy(2015)}]{2015SIoffeCSzegedy}
Sergey Ioffe and Christian Szegedy. 2015.
\newblock Batch normalization: Accelerating deep network training by reducing
  internal covariate shift.
\newblock In {\em Proc. ICML\/}.

\bibitem[{Irsoy and Cardie(2014)}]{irsoy2014deep}
Ozan Irsoy and Claire Cardie. 2014.
\newblock Deep recursive neural networks for compositionality in language.
\newblock In {\em Proc. NIPS\/}.

\bibitem[{Joulin and Mikolov(2015)}]{joulin2015inferring}
Armand Joulin and Tomas Mikolov. 2015.
\newblock Inferring algorithmic patterns with stack-augmented recurrent nets.
\newblock In {\em Proc. NIPS\/}.

\bibitem[{Kalchbrenner et~al.(2014)Kalchbrenner, Grefenstette, and
  Blunsom}]{kalchbrenner2014convolutional}
Nal Kalchbrenner, Edward Grefenstette, and Phil Blunsom. 2014.
\newblock A convolutional neural network for modelling sentences.
\newblock In {\em Proc. ACL\/}.

\bibitem[{Kiperwasser and Goldberg(2016)}]{kiperwasser2016easy}
Eliyahu Kiperwasser and Yoav Goldberg. 2016.
\newblock Easy-first dependency parsing with hierarchical tree {LSTM}s.
\newblock {arXiv}:1603.00375.

\bibitem[{Li et~al.(2015)Li, Luong, Jurafsky, and Hovy}]{li2015tree}
Jiwei Li, Minh-Thang Luong, Dan Jurafsky, and Eudard Hovy. 2015.
\newblock When are tree structures necessary for deep learning of
  representations?
\newblock In {\em Proc. EMNLP\/}.

\bibitem[{Mou et~al.(2016)Mou, Rui, Li, Xu, Zhang, Yan, and
  Jin}]{mou2015recognizing}
Lili Mou, Men Rui, Ge~Li, Yan Xu, Lu~Zhang, Rui Yan, and Zhi Jin. 2016.
\newblock Natural language inference by tree-based convolution and heuristic
  matching.
\newblock In {\em Proc. ACL\/}.

\bibitem[{Nivre(2003)}]{nivre2003efficient}
Joakim Nivre. 2003.
\newblock An efficient algorithm for projective dependency parsing.
\newblock In {\em Proc. IWPT\/}.

\bibitem[{Pennington et~al.(2014)Pennington, Socher, and
  Manning}]{pennington2014glove}
Jeffrey Pennington, Richard Socher, and Christopher~D. Manning. 2014.
\newblock Glo{V}e: Global vectors for word representation.
\newblock In {\em Proc. {EMNLP}\/}.

\bibitem[{Rockt{\"a}schel et~al.(2016)Rockt{\"a}schel, Grefenstette, Hermann,
  Ko{\v{c}}isk{\`y}, and Blunsom}]{rocktaschel2015reasoning}
Tim Rockt{\"a}schel, Edward Grefenstette, Karl Moritz Hermann, Tom{\'a}{\v{s}}
  Ko{\v{c}}isk{\`y}, and Phil Blunsom. 2016.
\newblock Reasoning about entailment with neural attention.
\newblock In {\em Proc. ICLR\/}.

\bibitem[{Shieber(1983)}]{shieber:1983:ACL}
Stuart~M. Shieber. 1983.
\newblock Sentence disambiguation by a shift-reduce parsing technique.
\newblock In {\em Proc. ACL\/}.

\bibitem[{Socher et~al.(2011{\natexlab{a}})Socher, Lin, Ng, and
  Manning}]{socher2011parsing}
Richard Socher, Cliff~C. Lin, Andrew~Y. Ng, and Christopher~D. Manning.
  2011{\natexlab{a}}.
\newblock Parsing natural scenes and natural language with recursive neural
  networks.
\newblock In {\em Proc. {ICML}\/}.

\bibitem[{Socher et~al.(2011{\natexlab{b}})Socher, Pennington, Huang, Ng, and
  Manning}]{socher2011semi}
Richard Socher, Jeffrey Pennington, Eric~H Huang, Andrew~Y. Ng, and
  Christopher~D. Manning. 2011{\natexlab{b}}.
\newblock Semi-supervised recursive autoencoders for predicting sentiment
  distributions.
\newblock In {\em Proc. EMNLP\/}.

\bibitem[{Srivastava et~al.(2014)Srivastava, Hinton, Krizhevsky, Sutskever, and
  Salakhutdinov}]{srivastava2014dropout}
Nitish Srivastava, Geoffrey Hinton, Alex Krizhevsky, Ilya Sutskever, and Ruslan
  Salakhutdinov. 2014.
\newblock Dropout: {A} simple way to prevent neural networks from overfitting.
\newblock {\em JMLR\/} 15.

\bibitem[{Sutskever et~al.(2014)Sutskever, Vinyals, and
  Le}]{sutskever2014sequence}
Ilya Sutskever, Oriol Vinyals, and Quoc~V. Le. 2014.
\newblock Sequence to sequence learning with neural networks.
\newblock In {\em Proc. {NIPS}\/}.

\bibitem[{Tai et~al.(2015)Tai, Socher, and Manning}]{tai2015improved}
Kai~Sheng Tai, Richard Socher, and Christopher~D. Manning. 2015.
\newblock Improved semantic representations from tree-structured long
  short-term memory networks.
\newblock In {\em Proc. ACL\/}.

\bibitem[{{Theano Development Team}(2016)}]{theano}
{Theano Development Team}. 2016.
\newblock {Theano: A {Python} framework for fast computation of mathematical
  expressions}.
\newblock {arXiv}:1605.02688.

\bibitem[{Tieleman and Hinton(2012)}]{tieleman2012lecture}
Tijmen Tieleman and Geoffrey Hinton. 2012.
\newblock Lecture 6.5 -- {RMSP}rop: Divide the gradient by a running average of
  its recent magnitude.
\newblock In {\em Neural Networks for Machine Learning\/}, Coursera.

\bibitem[{Titov and Henderson(2010)}]{titov2010latent}
Ivan Titov and James Henderson. 2010.
\newblock A latent variable model for generative dependency parsing.
\newblock In Harry Bunt, Paola Merlo, and Joakim Nivre, editors, {\em Trends in
  Parsing Technology\/}, Springer.

\bibitem[{Vendrov et~al.(2016)Vendrov, Kiros, Fidler, and
  Urtasun}]{DBLP:journals/corr/VendrovKFU15}
Ivan Vendrov, Ryan Kiros, Sanja Fidler, and Raquel Urtasun. 2016.
\newblock Order-embeddings of images and language.
\newblock In {\em Proc. ICLR\/}.

\bibitem[{Wang and Jiang(2016)}]{wang2015learning}
Shuohang Wang and Jing Jiang. 2016.
\newblock Learning natural language inference with {LSTM}.
\newblock In {\em Proc. NAACL\/}.

\bibitem[{Zhang et~al.(2015)Zhang, Zhao, and
  LeCun}]{DBLP:journals/corr/ZhangZL15}
Xiang Zhang, Junbo Zhao, and Yann LeCun. 2015.
\newblock Character-level convolutional networks for text classification.
\newblock In {\em Proc. NIPS\/}.

\bibitem[{Zhang et~al.(2016)Zhang, Lu, and Lapata}]{zhang2016top}
Xingxing Zhang, Liang Lu, and Mirella Lapata. 2016.
\newblock Top-down tree long short-term memory networks.
\newblock In {\em Proc. NAACL\/}.

\end{thebibliography}

\end{document}